\documentclass[a4paper]{styles/svproc}
\usepackage{url}
\usepackage{cite}
\usepackage{amsmath,amssymb,amsfonts}
\usepackage{algorithm}
\usepackage{algorithmic}
\usepackage{graphicx}
\usepackage{textcomp}
\usepackage{xcolor}
\usepackage{soul}
\usepackage{booktabs}
\usepackage{tabularx}
\usepackage{multirow}
\usepackage{subfigure}
\usepackage{makecell}
\usepackage{array}

\begin{document}
\mainmatter              
 \title{FlipItRight: Stable Pose-Targeted Throw-Flip Across Diverse Objects}
\titlerunning{Stable Pose-Targeted Throw-Flip Across Diverse Objects}
%
\author{Axel Dawne \and Shinkyu Park}
\authorrunning{Axel Dawne et al.} 
%
\tocauthor{Axel Dawne, Shinkyu Park}
\institute{King Abdullah University of Science and Technology (KAUST), Saudi Arabia\\
\email{axel.aribowo@kaust.edu.sa, shinkyu.park@kaust.edu.sa}
}

\maketitle              

\begin{abstract}
We propose \textit{FlipItRight}, a framework for stable planar pose-targeted throw-flip with a high-DoF manipulator. The task is decomposed into an object-level planner, which generates candidate release states satisfying the desired landing pose, and a robot-level planner, which evaluates executability and constructs a feasible swing motion. Treating the release state as an explicit intermediate representation enables principled candidate filtering, adaptive selection of release and pre-swing configurations, and structured near-release motion design---in particular, approximately constant end-effector velocities during the final swing phase to improve robustness to release-timing uncertainty. We validate on a real platform across objects of varying shape, size, and mass, achieving a 90\% success rate across $120$ trials. Ablation studies confirm that each design choice contributes to throwing performance, and the framework requires no prior data or learned model, enabling direct deployment on new objects and targets without environment-specific calibration or data collection.

\end{abstract}

\section{Introduction}
\vspace{-1.5em}
\begin{figure}
    \centering
    \includegraphics[width=0.8\textwidth]{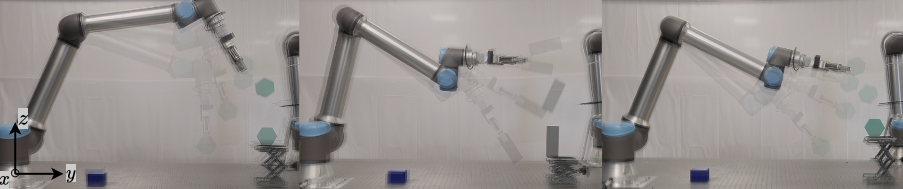}
    \caption{FlipItRight across different throw-flip scenarios.  For different objects and target landing poses, the framework generates scenario-specific throwing motions rather than relying on fixed initial or release robot configurations.}
    \label{fig:main_image}
    \vspace{-1em}
\end{figure}

Robotic throwing is one of the most demanding manipulation tasks: once an object leaves the gripper, the robot loses all direct control, and the subsequent trajectory is determined entirely by the release state---the position, orientation, and velocities at the moment of release. While considerable progress has been made in recent years~\cite{ref-1_dof_pitching,ref-control_throwing_one_joint_robot,ref-parts_assembly,ref-stochastic_dyn_throwing,ref-dual_throw_toss,ref-high_speed_throwing_kinetic_chain,ref-motion_ball_throwing,ref-optimal_control_manifold,ref-planning_impact_driven,ref-sampling_based,ref-throwing_motion_generation_using_opt,ref-bimanual_dynamic_grab_toss,ref-ball_throwing_visual_feedback,ref-adaptive_mobile_manipulator_throwing,ref-dartbot,ref-deep-predictive_policy_throwing,ref-dynamic_throwing_material_handling,ref-learning_throwing_catching,ref-learning_throw_decision_transformers,ref-learning-throw-flip,ref-task_level_throwing,ref-throwing_objects_moving_basket,ref-tossingbot}, most existing methods focus exclusively on achieving accurate \emph{landing position}. Achieving \emph{pose-targeted throw-flip}---controlling both landing position and orientation simultaneously---is significantly harder:  the release state must regulate not only the linear position and velocity needed for accurate landing position, but also the angular position and angular velocity needed to control landing orientation, since each contributes to both the landing position and orientation in a coupled manner. For high-DoF manipulators, this is further compounded by tight coupling between end-effector motion and joint configuration, making it a high-dimensional, tightly constrained planning problem. Precise landing pose matters in logistics automation, where items must arrive in a target orientation for downstream processing, and in assembly, where a part must land face-up before a subsequent operation.

A natural approach is to search directly over robot joint trajectories for one that achieves the desired release state. However, because both the release-state search and swing-motion planning are non-convex and sensitive to initialization, such a joint search is difficult to make reliable across geometrically diverse objects without a principled intermediate representation to guide it. Systematic experimental evaluation of pose-targeted throw-flip across diverse objects remains scarce, leaving the practical capabilities of high-DoF manipulators in this setting largely uncharacterized.

This work addresses that gap directly. As shown in Fig.~\ref{fig:main_image}, we study the planar setting: the throwing motion confined to the \(yz\)-plane and the object free to rotate about the \(x\)-axis during flight. Our main contribution is the design and real-robot validation of \textit{FlipItRight}, a planner for planar pose-targeted throw-flip evaluated across objects spanning a broad range of shapes, sizes, and masses. 

Specifically, we contribute: \textit{(i)} We study stable planar throw-flip to a prescribed landing configuration, requiring the object to settle on the intended support face, at the correct base height, and in the desired orientation. To our knowledge, this is the first real-robot study to systematically evaluate pose-targeted throw-flip across objects of varying shapes, sizes, and masses, and across multiple target poses. \textit{(ii)} We introduce the object release state as an explicit intermediate representation linking landing-pose requirements to robot motion generation, and propose a two-stage planning framework: an object-level  planner generates candidate release states consistent with the desired landing pose, and a robot-level planner assesses their executability, selects a feasible candidate, and constructs a swing motion that realizes it. \textit{(iii)} The explicit release-state representation enables principled candidate filtering, adaptive selection of release and pre-swing configurations, and structured near-release motion design---in particular, imposing approximately constant end-effector velocities during the final swing phase to improve robustness to release-timing uncertainty. Ablation studies confirm that each design choice contributes to throwing performance.

The paper is organized as follows: Section~\ref{sec:related_work} surveys related work. Section~\ref{sec:preliminaries} formulates the problem. Section~\ref{sec:flipitright} describes the \textit{FlipItRight} planning framework in detail. Section~\ref{sec:experiments} presents real-robot experiments across four objects and three target distances each, along with ablation studies. Section~\ref{sec:conclusions} concludes.

\section{Related Work} \label{sec:related_work}
Prior work on robotic throwing can be organized along two axes: the task objective and the scope of objects considered. Most methods targeting a single object focus on landing position~\cite{ref-control_throwing_one_joint_robot, ref-stochastic_dyn_throwing, ref-dual_throw_toss, ref-sampling_based, ref-throwing_motion_generation_using_opt, ref-bimanual_dynamic_grab_toss, ref-ball_throwing_visual_feedback, ref-adaptive_mobile_manipulator_throwing, ref-deep-predictive_policy_throwing, ref-dynamic_throwing_material_handling, ref-learning_throwing_catching, ref-task_level_throwing}, with only one addressing full landing pose~\cite{ref-parts_assembly}; a further set studies release state design without evaluating landing outcomes directly~\cite{ref-1_dof_pitching, ref-high_speed_throwing_kinetic_chain, ref-motion_ball_throwing}. Among methods evaluated over a set of objects, the majority again target landing position~\cite{ref-dartbot, ref-learning_throw_decision_transformers, ref-throwing_objects_moving_basket, ref-tossingbot}, while a smaller body of work addresses landing pose~\cite{ref-optimal_control_manifold, ref-planning_impact_driven, ref-tossnet, ref-learning-throw-flip}. Our work falls in this last category and, to the best of our knowledge, is the first to systematically evaluate planar landing pose across objects of varying shapes, sizes, and masses.

Existing methods can be viewed as either \emph{model-based} or \emph{learning-based}. Model-based approaches typically rely on analytical models, trajectory optimization, or constrained search over feasible throwing motions~\cite{ref-stochastic_dyn_throwing,ref-high_speed_throwing_kinetic_chain,ref-optimal_control_manifold,ref-planning_impact_driven}. Learning-based approaches instead use data to compensate for modeling errors or to directly predict throwing parameters and outcomes, often improving robustness in less structured settings~\cite{ref-tossingbot,ref-tossnet,ref-learning-throw-flip, ref-learning_throw_decision_transformers}. Our approach is model-based, relying on analytical flight dynamics and trajectory optimization.

Another important distinction is between \emph{non-prehensile} and \emph{prehensile} throwing. Non-prehensile methods generally assume the object is not rigidly constrained during the swing phase, for example resting on a flat end-effector surface~\cite{ref-optimal_control_manifold,ref-parts_assembly}. Prehensile methods maintain grasp until release and therefore require coordinating object release with the robot swing trajectory~\cite{ref-planning_impact_driven,ref-tossnet,ref-learning-throw-flip}. This makes prehensile pose-targeted throwing particularly challenging, since release contact effects, including timing, friction, and non-instantaneous detachment, can also affect the object's subsequent flight trajectory.

We now turn to the pose-targeted throwing methods most closely related to our setting. TossNet~\cite{ref-tossnet} jointly learns robot throwing dynamics and object flight trajectories for arbitrary rigid objects, then exploits a bisection search over the learned forward model to find toss actions that land a ball in a target container. However, in the demonstrated task, the object motion is restricted to linear displacement without rotation.

The work of \cite{ref-learning-throw-flip} introduces a throw-flip primitive based on impulse--momentum principles, parameterized by pitch, speed, and braking damping, and exploits a \emph{temporal hinge} at the moment of gripper release. Local models map these parameters to landing outcomes either directly or via a predicted detach state coupled with ballistic dynamics; hardware transfer is validated across objects with varying centers of mass. Their evaluation adopts a thresholded success criterion with tolerances of $\pm 5\,\mathrm{cm}$ and $\pm 45^\circ$—coarser than the explicit stable-landing criterion we employ. Notably, \cite{ref-learning-throw-flip} relies on an impedance controller, which requires direct torque actuation available on their Franka Emika Panda platform but not on standard position/velocity-controlled manipulators such as the UR10e used in our work. Furthermore, even accounting for the Velcro-assisted landing surface, our mean landing-distance errors fall well within the ±5 cm tolerance across most objects and scenarios, suggesting that competitive or superior position accuracy is achievable under standard joint trajectory control without iterative data collection.

Zermane et al.~\cite{ref-planning_impact_driven} present a common planning interface for three impact-driven logistics skills---tossing, grabbing, and boxing---by mapping task-space landing and impact specifications to a joint-space via-point, which a kinodynamic planner then connects to the current robot state. Although landing position, orientation, and impact velocity are included, real-robot tossing experiments focus on container targets and impact-velocity variation without reporting systematic pose accuracy or repeatability.

Our method distinguishes itself from these approaches by treating the object release state as an explicit intermediate representation that bridges landing-pose requirements and robot motion generation. Rather than searching directly over full robot trajectories or relying on a fixed throwing primitive, this representation lets the planner screen landing-valid release candidates for kinematic executability, adaptively select the swing's release and initial postures, and impose structured near-release motion constraints prior to execution. We validate this design on a real robot using explicit landing-pose metrics and a strict stable-landing success criterion across geometrically and physically diverse objects.

\section{Preliminaries and Problem Description}
\label{sec:preliminaries}
We design a trajectory planning pipeline that determines the required release state---the position, orientation, and velocities at which the object must leave the gripper---and computes a feasible robot swing motion that achieves this state at the designated release instant. 

Let 
\(\boldsymbol q(t) := (q_{\mathrm{lift}}(t), \, q_{\mathrm{elbow}}(t), \, q_{\mathrm{wrist}}(t)) \in \mathbb{R}^3\)
denote the actuated joint angles, and let \(\dot{\boldsymbol q}(t)\) denote the corresponding joint velocities. The end-effector (EE) pose is obtained from the forward-kinematics map
\((\boldsymbol p(t),\theta(t)) = f_{\mathrm{FK}}(\boldsymbol q(t))\), where \(\boldsymbol p(t)\in\mathbb{R}^2\) is the EE position in the \(yz\)-plane, and \(\theta(t)\in\mathbb{R}\) is the orientation angle about the \(x\)-axis, following the right-hand rule; see Fig.~\ref{fig:main_image}. The corresponding EE linear and angular velocities are computed from the differential kinematics
\((\boldsymbol v(t),\omega(t)) = J_{\mathrm{ee}}(\boldsymbol q(t))\dot{\boldsymbol q}(t)\), where \(\boldsymbol v(t)=(v_y(t),v_z(t))\in\mathbb{R}^2\) is the EE linear velocity in the \(yz\)-plane, \(\omega(t)\in\mathbb{R}\) is the angular velocity about the \(x\)-axis, and \(J_{\mathrm{ee}}\) is the corresponding geometric Jacobian.

We collect these quantities into the end-effector state
\begin{equation}
\label{eq:ee_state}
\boldsymbol{\xi}_{\mathrm{ee}}(t)
\triangleq
\left( \boldsymbol p(t), \, \theta(t), \, \boldsymbol v(t), \, \omega(t) \right)
=
f_{\mathrm{kine}}\!\left(\boldsymbol q(t),\dot{\boldsymbol q}(t)\right),
\end{equation}
where \(f_{\mathrm{kine}}\) denotes the combined pose and velocity kinematic mapping.
Conversely, inverse kinematics maps a desired EE pose back to a joint configuration. For a desired pose \((\boldsymbol p,\theta)\), an inverse-kinematics solution is any joint configuration $\boldsymbol q$ belonging to
\begin{equation}
\label{eq:ik}
\left\{
\tilde{\boldsymbol q} \in \mathbb R^3 \,\big|\,
(\boldsymbol p, \, \theta) = f_{\mathrm{FK}}(\tilde{\boldsymbol q})
\right\}.
\end{equation}

The desired landing pose comprises a target position \(\boldsymbol p_{\text{des}} \in \mathbb R^2\) in the $yz$-plane and a target orientation \(\theta_{\text{des}}\) about the $x$-axis. We make the following assumptions in our problem formulation: (i) \textbf{Rigid grasp:} the object moves rigidly with the end effector throughout the swing, with no slip at the gripper-object interface. (ii) \textbf{Decoupled load dynamics:} object inertia does not perturb the manipulator, so the robot is modeled independently of the grasped load. (iii) \textbf{Negligible aerodynamics:} air resistance is neglected during flight. (iv) \textbf{Instantaneous release:} the object detaches at a single instant \(t=T_{\mathrm{end}}\) and subsequently undergoes ballistic free flight under gravity alone.

While the first three assumptions are standard in model-based throwing and hold to a good approximation in our setting, the instantaneous-release assumption is a deliberate planning convenience: the framework compensates for the known violation of this assumption through gripper command pre-timing, as described in Section~\ref{sec:experiments}.

\section{FlipItRight: Two-Stage Planning for Stable Throw-Flip} \label{sec:flipitright}

\begin{figure}[t]
    \centering
    \includegraphics[width=.8\textwidth]{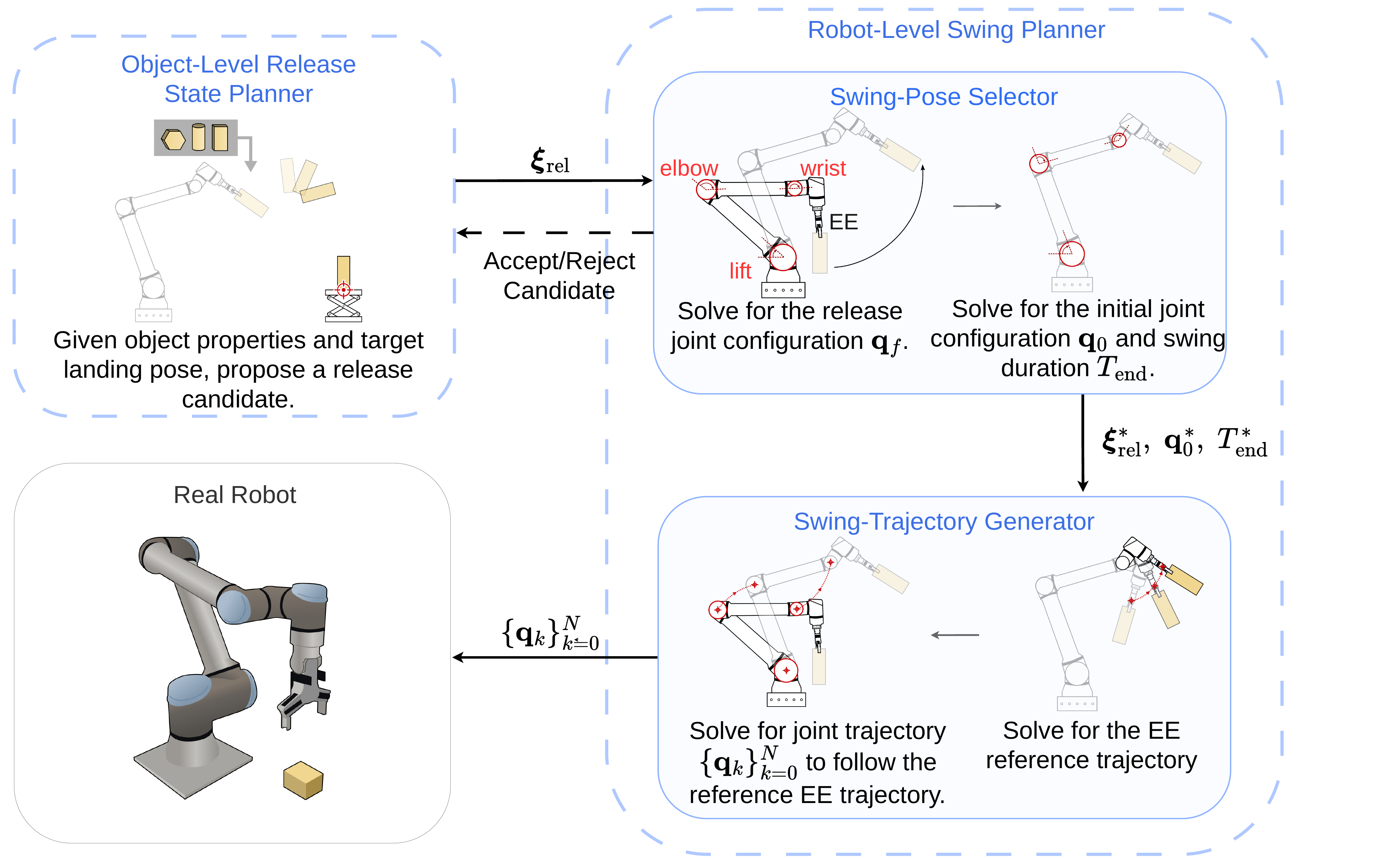}
    \vspace{-0.5em}
    \caption{Proposed pose-targeted throw-flip framework. The object-level planner proposes a release state \(\boldsymbol{\xi}_{\mathrm{rel}}\); the robot-level planner validates it and computes swing variables; the swing-pose selector solves for the release configuration \(\mathbf{q}_f\), initial configuration \(\mathbf{q}_0\), and swing duration \(T_{\mathrm{end}}\); and the swing-trajectory generator produces the joint trajectory \(\{\mathbf{q}_k\}_{k=0}^{N}\) for execution.}
    \label{fig:framework_overview}
    \vspace{-1em}
\end{figure}

We present a pose-targeted throwing framework comprising two sequentially coupled modules: the \textbf{object-level release state planner} and the \textbf{robot-level swing motion planner}, as illustrated in Fig.~\ref{fig:framework_overview}. The key idea is to treat the release state as an intermediate representation between landing-pose requirements and robot motion generation. Since the landing pose depends only on the object state at release, while the robot's role is to realize a physically executable release state, the proposed decomposition separates object-level landing feasibility from robot-level executability. The object-level planner generates candidate release states, which the robot-level planner evaluates for kinematic feasibility and realizes through a swing motion. Candidates that are infeasible are rejected and replaced before trajectory generation. This feasibility-filtering mechanism is particularly effective because both the release-state search and swing-motion planning problems are non-convex and sensitive to initialization, allowing computational effort to be focused on executable solutions while enabling adaptive use of robot kinematics and structured motion design around the release event.

\subsection{Object-Level Release State Planner} \label{subsec:object_release_planner}

The object release planner computes a candidate object release state $\boldsymbol{\xi}_{\text{rel}} \;\triangleq\; (\boldsymbol p_{\text{rel}},\, \theta_{\text{rel}},\, \boldsymbol v_{\text{rel}},\, \omega_{\text{rel}} )$ at $t = T_{\mathrm{end}}$---comprising release position $\boldsymbol p_{\text{rel}}$, orientation $\theta_{\text{rel}}$, linear velocity $\boldsymbol v_{\text{rel}}$, and angular velocity $\omega_{\text{rel}}$---such that an object released precisely at this state lands at the desired pose after free flight.
We compute $\boldsymbol{\xi}_{\mathrm{rel}}$ by solving the constrained nonlinear optimization:
\begin{equation}
\label{eq:opt_rel}
\begin{aligned}
\textstyle\min_{\underline{\boldsymbol{\xi}}_{\mathrm{rel}} \;\le\; \boldsymbol{\xi}_{\mathrm{rel}} \;\le\; \overline{\boldsymbol{\xi}}_{\mathrm{rel}}}\quad
& \lambda_y \big(y_{\text{des}}-y_{\text{landing}}\big)^{2}
\;+\; \lambda_\theta \big(\theta_{\text{des}}-\theta_{\text{landing}}\big)^{2} \\
&\textstyle \quad + \tfrac{\lambda_{\omega}}{2}I\,\omega_{\text{landing}}^{2}
+\tfrac{\lambda_{v}}{2}m\,\|\boldsymbol v_{\text{landing}}\|_2^{2}
+ \lambda_g m g z_{0} \\[3pt]
\text{s.t.}\quad
& \big(\boldsymbol p_{\text{landing}},\,\boldsymbol v_{\text{landing}},
\,\omega_{\text{landing}},\,\theta_{\text{landing}},\,T_{\text{flight}}\big)
=
\mathcal{B}\!\left(\boldsymbol{\xi}_{\mathrm{rel}}\right).
\end{aligned}
\end{equation}
Here, $\underline{\boldsymbol{\xi}}_{\mathrm{rel}}$ and $\overline{\boldsymbol{\xi}}_{\mathrm{rel}}$ are componentwise lower and upper bounds defining the search region.
The horizontal release-position bound is set relative to the desired landing position, constraining the release point to lie $10$--$15$ cm before the target to keep flight distances comparable and avoid unnecessarily long throws; the release-height and velocity bounds limit excessive landing energy. The constraint maps a candidate $\boldsymbol{\xi}_{\mathrm{rel}}$ to the corresponding landing state---position $\boldsymbol p_{\text{landing}}$, orientation $\theta_{\text{landing}}$, linear velocity $\boldsymbol v_{\text{landing}}$, angular velocity $\omega_{\text{landing}}$, and flight time $T_{\text{flight}}$---
via the ballistic propagation operator $\mathcal B$, defined in Algorithm~\ref{alg:time_of_flight}. 

\begin{algorithm} [t]
\caption{Geometry-Aware Ballistic Propagation}
\label{alg:time_of_flight}
\begin{algorithmic}[1]
\REQUIRE Release state $\boldsymbol{\xi}_{\mathrm{rel}}$, object geometry $\mathcal G$, destination height $z_{\mathrm{des}}$, time step $\Delta t$
\STATE Extract $(y_0,z_0)$, $(v_{y0},v_{z0})$, $\theta_{0}$, and $\omega_{0}$ from $\boldsymbol{\xi}_{\mathrm{rel}}$
\STATE Assign $(y,z)\!\gets\!(y_0,z_0)$;\; $(v_y,v_z)\!\gets\!(v_{y0},v_{z0})$;\; $\theta\!\gets\!\theta_{0}$;\; $\omega\!\gets\!\omega_{0}$;\; $T\!\gets\!0$
\STATE Compute initial lowest vertex height $z_{\text{min}}$ from $\mathcal G$ at orientation $\theta$
\WHILE{$z_{\min} > z_{\mathrm{des}}$}
    \STATE $y\!\gets\!y+v_y\Delta t$;\;
           $z\!\gets\!z+v_z\Delta t-\tfrac{1}{2}g\Delta t^2$;\;
    \STATE $\theta\!\gets\!\theta+\omega\Delta t$;\; $T\!\gets\!T+\Delta t$
    \STATE Update $z_{\text{min}}$ from $\mathcal G$ at orientation $\theta$
\ENDWHILE
\STATE \textbf{return} $T,\; (y,z),\; (v_y,v_z),\; \omega,\; \theta$
\end{algorithmic}
\end{algorithm}

The objective penalizes landing-position error (weight \(\lambda_y=1\)), landing-orientation error (weight \(\lambda_\theta=1\)), and a regularizer consisting of rotational kinetic, translational kinetic, and gravitational potential energy with weights \(\lambda_\omega=0.003\), \(\lambda_v=0.001\), and \(\lambda_g=0.001\), respectively, where \(z_0\) denotes the vertical height of the object's center of mass at release. The energy regularizer discourages high-energy release states, which tend to produce large landing velocities and reduce the likelihood of a stable landing. Under the assumptions in Section~\ref{sec:preliminaries}, angular velocity is constant during flight, i.e., $\omega_{\text{landing}}=\omega_{\text{rel}}$.

The problem is solved via SLSQP~\cite{ref-numerical_opt}. Since~\eqref{eq:opt_rel} is non-convex, we adopt a random-restart strategy, sampling the initial guess uniformly over the bounding box,
\(
\boldsymbol{\xi}_{0}\sim\mathrm{Unif}([\underline{\boldsymbol{\xi}}_{\mathrm{rel}},\,\overline{\boldsymbol{\xi}}_{\mathrm{rel}}]),
\)
re-solving up to \(N_{\mathrm{obj}}=20\) times. Each candidate is retained only if it satisfies the object-level thresholds $|y_{\mathrm{landing}}-y_{\mathrm{des}}| \leq \epsilon_y^{\mathrm{obj}}$ and $|\theta_{\mathrm{landing}}-\theta_{\mathrm{des}}| \leq \epsilon_\theta^{\mathrm{obj}},$
with \(\epsilon_y^{\mathrm{obj}}=0.005~\mathrm{m}\) and \(\epsilon_\theta^{\mathrm{obj}}=0.5^\circ\). Candidates satisfying these thresholds are forwarded to the robot-level swing planner; otherwise, the candidate is rejected and the overall swing-planner loop invokes the object-level planner again to sample and optimize a new release candidate.

\subsection{Robot-Level Swing Motion Planner} \label{subsec:robot_swing_planner}
Given a candidate release state \(\boldsymbol{\xi}_{\mathrm{rel}}\), the \textit{Swing-Posture Selector} checks feasibility by computing a compatible robot release posture, initial posture, and swing duration. If feasible, the \textit{Swing-Trajectory Generator} computes the swing motion.

\subsubsection{Swing-Posture Selector.} \label{subsubsec: initial_joint_config_optimizer}
Since the object-level planner outputs a release state \(\boldsymbol{\xi}_{\mathrm{rel}}\) in object-frame coordinates, we use a fixed rigid-grasp map \(\mathcal{M}(\cdot)\) to convert between the end-effector state $\boldsymbol{\xi}_{\mathrm{ee}}$ and object center-of-mass state $\boldsymbol{\xi}_{\mathrm{obj}}$: \(
\boldsymbol{\xi}_{\mathrm{obj}}(t)
=
\mathcal{M}\!\left(\boldsymbol{\xi}_{\mathrm{ee}}(t)\right)
\), so that the terminal condition $\boldsymbol{\xi}_{\mathrm{obj}}(T_{\mathrm{end}}) = \boldsymbol{\xi}_{\mathrm{rel}}$ can be enforced through the end-effector trajectory. We first compute the release joint configuration \(\boldsymbol q_f\) via inverse-kinematics \eqref{eq:ik}, requiring $\mathcal M (\boldsymbol{\xi}_{\mathrm{ee}}(T_{\mathrm{end}}))$ to match the pose component $(\boldsymbol p_{\mathrm{rel}},\theta_{\mathrm{rel}})$ of \(\boldsymbol{\xi}_{\mathrm{rel}}\).

The swing-posture selector then solves for the remaining swing variables under two design requirements. First, to improve robustness to timing errors and non-instantaneous gripper opening, we require the end-effector velocity to have converged to its target over a stabilization interval \([T_{\text{stab}},\,T_{\mathrm{end}}]\), where $T_{\mathrm{stab}} = \alpha T_{\mathrm{end}}, \, 0 < \alpha < 1$.\footnote{Precisely matching the desired velocity at a single instant $T_{\mathrm{end}}$ is highly sensitive to timing errors.} Second, the resulting joint-velocity profile must respect joint velocity limits. These two requirements naturally induce a two-stage profile: joint velocity ramps linearly from \(\boldsymbol 0\) to \(\dot{\boldsymbol q}_{\text{stab}}\) over $[0,\,T_{\text{stab}}]$, then the end-effector velocity remains approximately constant over \([T_{\text{stab}},\,T_{\mathrm{end}}]\).

We also define
\(
T_{\mathrm{mid}} = \frac{T_{\mathrm{stab}} + T_{\mathrm{end}}}{2}
\)
as an intermediate evaluation time used in the velocity-constancy residual to verify that the velocity remains approximately constant throughout the stabilization interval, not only at its endpoints.

Together, these requirements lead to a bounded nonlinear least-squares problem over 
\(\boldsymbol z = (\boldsymbol q_{0},\, \dot{\boldsymbol q}_{\text{stab}},\, \dot{\boldsymbol q}_{\mathrm{end}},\, T_{\mathrm{end}}) \in \mathbb{R}^{10}\), where $\boldsymbol q_{0} \in \mathbb R^3$ is the initial joint configuration, \(\dot{\boldsymbol q}_{\text{stab}}, \dot{\boldsymbol q}_{\mathrm{end}} \in \mathbb R^3\) are the joint velocities at \(t=T_{\text{stab}}\) and \(t=T_{\mathrm{end}}\), respectively, and \(T_{\mathrm{end}} \in \mathbb{R}\) is the swing duration:

\begin{align} \label{eq:initial_joint_configuration_optimizer}
& \textstyle\min_{\boldsymbol z = (\boldsymbol q_{0},\,
      \dot{\boldsymbol q}_{\text{stab}},\,
      \dot{\boldsymbol q}_{\mathrm{end}},\,
      T_{\mathrm{end}})
     }
\quad
 \| \boldsymbol r_{\mathrm{rel}} \|^2 + \frac{w_{\text{const}}}{s_{\mathrm{const}}} \|\boldsymbol r_{\mathrm{const}}\|^2 + w_{\text{acc}} \|\boldsymbol a_{1}\|^2 + w_{\text{acc}} \|\boldsymbol a_{2}\|^2 \nonumber \\ 
& \textstyle\qquad\qquad\qquad\qquad\qquad\qquad +  \frac{w_{\text{swing}}}{s_{\mathrm{swing}}} \|\boldsymbol r_{\text{swing}}\|^2 + \frac{w_{\text{height}}}{s_{\mathrm{height}}}\| r_{\text{height}} \|^2
\\[6pt]
&\begin{aligned}
\text{s.t.}\quad
&
\dot{\boldsymbol q}_{\min}
\le
\dot{\boldsymbol q}_{\text{stab}},
\dot{\boldsymbol q}_{\mathrm{end}}
\le
\dot{\boldsymbol q}_{\max}, ~
T_{\min}
\le
T_{\mathrm{end}}
\le
T_{\max}, ~
\boldsymbol q_{0}^{\min}
\le
\boldsymbol q_{0}
\le
\boldsymbol q_{0}^{\max}. \nonumber
\end{aligned}
\end{align}

Recall \(\boldsymbol{\xi}_{\mathrm{rel}} = (\boldsymbol p_{\mathrm{rel}}, \theta_{\mathrm{rel}}, \boldsymbol v_{\mathrm{rel}}, \boldsymbol \omega_{\mathrm{rel}})\). Writing \((\boldsymbol p_t,\, \theta_t, \, \boldsymbol v_t, \, \omega_t) = \boldsymbol{\xi}_{\mathrm{obj}}(t) = \mathcal M(\boldsymbol{\xi}_{\mathrm{ee}}(t)) \) for the achieved object state at time $t$, we evaluate these quantities at \(t \in \{0, T_{\mathrm{stab}}, T_{\mathrm{mid}}, T_{\mathrm{end}}\}\). The resulting terms are:

\begin{itemize}

\item \textbf{Release-state residual \(\boldsymbol r_{\mathrm{rel}}\):}
Penalizes normalized deviations between the achieved and candidate release state in position, orientation, and velocity: ${\boldsymbol r}_{\mathrm{rel}}
=
\big(
(\boldsymbol p_{T_{\mathrm{end}}} - \boldsymbol p_{\mathrm{rel}})/s_p, 
\mathrm{LogRot}\!\big(R_{\mathrm{rel}} R(\boldsymbol \eta_{T_{\mathrm{end}}})^{-1}\big)/s_R, 
(\boldsymbol v_{T_{\text{stab}}} - \boldsymbol v_{\mathrm{rel}})/s_v,
(\boldsymbol \omega_{T_{\text{stab}}} - \boldsymbol \omega_{\mathrm{rel}})/s_\omega
\big).$
Here, \(\boldsymbol \eta_{T_{\mathrm{end}}}\) denotes the achieved orientation quaternion at release, and \(R(\boldsymbol \eta_{T_{\mathrm{end}}})\) is its corresponding rotation matrix. Similarly, \(R_{\mathrm{rel}}\) is the rotation matrix associated with the target release orientation \(\theta_{\mathrm{rel}}\). 

\item \textbf{Velocity-constancy residual \(\boldsymbol r_{\mathrm{const}}\):}
Encourages a steady end-effector velocity over the interval \([T_{\text{stab}},\,T_{\mathrm{end}}]\) by penalizing changes in linear and angular velocity: $\boldsymbol r_{\mathrm{const}}
=
(
\boldsymbol v_{T_{\text{mid}}} - \boldsymbol v_{T_{\text{stab}}}, 
\boldsymbol \omega_{T_{\text{mid}}} - \boldsymbol \omega_{T_{\text{stab}}},
\boldsymbol v_{T_{\mathrm{end}}} - \boldsymbol v_{T_{\text{stab}}},
\boldsymbol \omega_{T_{\mathrm{end}}} - \boldsymbol \omega_{T_{\text{stab}}}
)$.

\item \textbf{Joint-acceleration residuals \(\boldsymbol a_1\) and \(\boldsymbol a_2\):}
Approximate the joint accelerations induced by the two-stage velocity profile and penalize aggressive joint-space motions: $\boldsymbol a_1
= \frac{\dot{\boldsymbol q}_{\text{stab}}}{T_{\text{stab}}}$ and $\boldsymbol a_2
= 
\frac{\dot{\boldsymbol q}_{\mathrm{end}} - \dot{\boldsymbol q}_{\text{stab}}}{T_{\mathrm{end}} - T_{\text{stab}}}.$

\item \textbf{Swing-direction residual \(\boldsymbol r_{\text{swing}}\) and height residual \(r_{\text{height}}\):}
Encourages outward and upward end-effector motion and penalize initial postures too close to the ground: $\boldsymbol r_{\text{swing}}
=
(
\max\{0, \varepsilon_y - (p_{T_{\mathrm{end}},y} - p_{0,y})\}, 
\max\{0, \varepsilon_z - (p_{T_{\mathrm{end}},z} - p_{0,z})\}
)$ and $
r_{\text{height}}
= \max\{0, z_{\min} - p_{0,z}\}$, where \(\varepsilon_y,\varepsilon_z > 0\) enforce the desired swing direction and \(z_{\min}\) is the required minimum starting height.

\item \textbf{Bounds on \(\boldsymbol q_0\):} The search region for the initial configuration is centered \(\boldsymbol q_f\), with radius determined by the reachable-time factor
\(
\Delta t_{\mathrm{reach}}
=
\left(1-\frac{\alpha}{2}\right)T_{\max},
\)
which approximates the maximum effective time available for joint displacement under the two-stage velocity profile: $\boldsymbol q_0^{\min} = \boldsymbol q_f - \dot{\boldsymbol q}_{\max}\Delta t_{\mathrm{reach}}$ and $\boldsymbol q_0^{\max} = \boldsymbol q_f + \dot{\boldsymbol q}_{\max}\Delta t_{\mathrm{reach}}.$

\item \textbf{Bounds on \(T_{\mathrm{end}}\):} The swing duration is bounded in \([T_{\min}, T_{\max}]\) to keep execution time reasonable. The upper bound \(T_{\max}\) is also determines the joint-displacement search radius for \(\boldsymbol q_0\).

\end{itemize}

End-effector quantities above (position \(\boldsymbol p\), linear velocity \(\boldsymbol v\), angular velocity \(\boldsymbol \omega\), orientation $\theta$) are from the forward-kinematics \eqref{eq:ee_state}. The normalization constants are set to \(s_p=0.01\,\mathrm{m}\), \(s_R=0.0174\,\mathrm{rad}\), \(s_v=0.05\,\mathrm{m/s}\), \(s_\omega=0.05\,\mathrm{rad/s}\), \(s_{\mathrm{const}}=2.5\times 10^{-3}\), \(s_{\mathrm{swing}}=10^{-4}\), and \(s_{\mathrm{height}}=10^{-4}\). The scalar weights are set to \(w_{\mathrm{const}}=0.25\), \(w_{\mathrm{acc}}=0.05\), and \(w_{\mathrm{swing}}=w_{\mathrm{height}}=10.0\). These scales and weights make residuals with different physical units numerically comparable and reflect the relative tolerance expected for each release-state component.

We solve \eqref{eq:initial_joint_configuration_optimizer} as a bounded nonlinear least-squares problem. Let \(\mathcal{R}(\boldsymbol z;\boldsymbol{\xi}_{\mathrm{rel}})=[r_1,\ldots,r_m]^\top\) denote the stacked residual vector whose entries are obtained from the normalized and weighted residual blocks in \eqref{eq:initial_joint_configuration_optimizer}, including the release-state residual, velocity-constancy residual, joint-acceleration residuals, swing-direction residual, and height residual. For numerical robustness, we first apply a soft-\(\ell_1\) warm-start strategy to these residual terms. Specifically, instead of minimizing the initial quadratic cost directly, we first minimize \(\sum_i \rho(r_i)\), where \(\rho(r_i)=2(\sqrt{1+r_i^2}-1)\). The resulting solution is then used to initialize a second solve using the standard quadratic least-squares objective, \(\|\mathcal{R}(\boldsymbol z;\boldsymbol{\xi}_{\mathrm{rel}})\|_2^2\), which corresponds to \eqref{eq:initial_joint_configuration_optimizer}.

A candidate is rejected if numerical errors occur or the object state cannot be evaluated at any required configuration. Among valid candidates, one is marked successful if the position, orientation, linear-velocity, and angular-velocity errors at release all fall below \(\epsilon_{\mathrm{rel}}=0.15\). If no successful candidate is found within 50 trials, feasible candidates are ranked lexicographically by \eqref{eq:initial_joint_configuration_optimizer} and \eqref{eq:opt_rel}, and the best-ranked candidate is retained.

\subsubsection{Swing-Trajectory Generator.}\label{subsubsec:swing_traj_generator}
After the Swing-Posture Selector identifies a feasible release candidate \(\boldsymbol{\xi}_{\mathrm{rel}}^{*}\), the Swing-Trajectory Generator computes a reference trajectory for the grasped object state \(\boldsymbol{\xi}_{\mathrm{obj}}(t)\) from the optimized initial configuration \(\boldsymbol q_0^\ast\), with terminal condition \(
\boldsymbol{\xi}_{\mathrm{obj}}(T_{\mathrm{end}}^\ast)
=
\boldsymbol{\xi}_{\mathrm{rel}}^\ast.
\)

We parameterize the end-effector reference using a two-phase trajectory, as discussed in the Swing-Posture Selector. The first phase (\(0\le t\le T_{\mathrm{stab}}\)) uses a cubic Hermite spline; the second phase (\(T_{\mathrm{stab}}<t\le T_{\mathrm{end}}^\ast\)) follows constant-velocity motion to reach $\boldsymbol{\xi}_{\mathrm{rel}}^\ast$ at \(T_{\mathrm{end}}^\ast\). We discretize with step \(h=0.002\,\mathrm{s}\), \(t_k=kh\), \(T_{\mathrm{end}}^{*}=Nh\), and switching index \(N_{\mathrm{stab}}\) satisfying \(T_{\mathrm{stab}}=N_{\mathrm{stab}}h\).

At \(t=0\), the end-effector state is obtained from forward kinematics at the optimized initial configuration: \((\boldsymbol p_0,\,\theta_0,\,\boldsymbol v_0,\,\omega_0)=f_{\mathrm{kine}}(\boldsymbol q_0^{*},\boldsymbol 0)\). Let \(\Delta T=T_{\mathrm{end}}^\ast-T_{\mathrm{stab}}\). To ensure that the EE reaches the desired release state \(\boldsymbol{\xi}_{\mathrm{ee}}^\ast=(\,\boldsymbol p^{*},\,\theta^{*},\,\boldsymbol v^{*},\,\omega^{*}\,)\) at \(T_{\mathrm{end}}^\ast\), we define the pre-release position as \(\boldsymbol p_{\mathrm{stab}}=\boldsymbol p^\ast-\boldsymbol v^\ast\Delta T\). Let \(\sigma_k=t_k/T_{\mathrm{stab}}\) for \(k=0,\ldots,N_{\mathrm{stab}}\). The reference position and linear velocity over the first phase are then:
\begin{align*}
\textstyle\boldsymbol p^{\,r}(t_k)
&= h_{00}(\sigma_k)\,\boldsymbol p_0
 + h_{10}(\sigma_k)\,T_{\mathrm{stab}}\,\boldsymbol v_0 
 + h_{01}(\sigma_k)\,\boldsymbol p_{\mathrm{stab}}
 + h_{11}(\sigma_k)\,T_{\mathrm{stab}}\,\boldsymbol v^{*}, \\
\textstyle\boldsymbol v^{\,r}(t_k)
&= \frac{1}{T_{\mathrm{stab}}}\Big[
      h'_{00}(\sigma_k)\,\boldsymbol p_0
    + h'_{10}(\sigma_k)\,T_{\mathrm{stab}}\,\boldsymbol v_0 
    + h'_{01}(\sigma_k)\,\boldsymbol p_{\mathrm{stab}}
    + h'_{11}(\sigma_k)\,T_{\mathrm{stab}}\,\boldsymbol v^{*}
\Big].
\end{align*}
where \(h_{00}, h_{10}, h_{01}, h_{11}\) denote the standard cubic Hermite basis functions and \(h'_{00}, h'_{10}, h'_{01}, h'_{11}\) are their derivatives with respect to \(\sigma_k\).
For \(k=N_{\mathrm{stab}}+1,\ldots,N\), with \(\tau_k=t_k-T_{\mathrm{stab}}\), the reference follows: $\boldsymbol p^{\,r}(t_k) = \boldsymbol p_{\mathrm{stab}}+\boldsymbol v^\ast \tau_k$ and $\boldsymbol v^{\,r}(t_k) = \boldsymbol v^\ast.$

The orientation reference is constructed analogously, with \(\theta_{\mathrm{stab}}=\theta^\ast-\omega^\ast\Delta T\), and \(\boldsymbol\eta^r(t_k)\) denotes the corresponding unit-quaternion. Given these reference trajectories, we find a joint-space trajectory using an offline optimization scheme: for the discrete reference sequence \(\{\boldsymbol p^{\,r}(t_k),\, \theta^r(t_k)\}_{k=0}^{N}\), we solve \eqref{eq:ik}  at each waypoint to obtain a nominal joint trajectory \(\{\bar{\boldsymbol q}_k\}_{k=0}^{N}\), used to warm-start the subsequent nonlinear least-squares refinement. 

We optimize the joint-space trajectory \(\{\boldsymbol q_k\}_{k=0}^{N} \), with the initial configuration fixed to the optimized swing-start posture: \(\boldsymbol q_0=\boldsymbol q_0^\ast\).
Joint velocities are computed via finite differences: $\boldsymbol u_k \triangleq \frac{\boldsymbol q_{k+1}-\boldsymbol q_k}{h}, \, k=0,\ldots,N-1.$
Letting \(\boldsymbol\eta(t_k)\) and \(\boldsymbol\eta^r(t_k)\) denote the unit-quaternion representation of the EE orientation and its reference, respectively, the tracking residuals at $t_k$ are: $\boldsymbol e_p(t_k) = \boldsymbol p^{\,r}(t_k) - \boldsymbol p(t_k)$,
$\boldsymbol e_v(t_k) = \boldsymbol v^{\,r}(t_k) - \boldsymbol v(t_k)$,
$\boldsymbol e_\omega(t_k) = \boldsymbol \omega^{\,r}(t_k) - \boldsymbol \omega(t_k)$,
and \(\boldsymbol e_R(t_k)=
\mathrm{LogRot}\!\big(R(\boldsymbol\eta^r(t_k))R(\boldsymbol\eta(t_k))^{-1}\big)\),
where \(\boldsymbol p(t_k)\), \(\boldsymbol\eta(t_k)\), \(\boldsymbol v(t_k)\), and \(\boldsymbol \omega(t_k)\) are obtained from the forward and differential kinematics defined in~\eqref{eq:ee_state}, evaluated at \((\boldsymbol q_k,\boldsymbol u_k)\).

The optimized trajectory solves:
\begin{align*}
\label{eq:os_cost}
&\textstyle\min_{\{\boldsymbol q_k\}_{k=1}^{N}}
\sum_{k=1}^{N}
\Big(
w_p\|\boldsymbol e_p(t_k)\|^2
+ w_v\|\boldsymbol e_v(t_k)\|^2
+ w_\omega\|\boldsymbol e_\omega(t_k)\|^2
+ w_\theta\|\boldsymbol e_\theta(t_k)\|^2
\Big)
\nonumber\\
&\textstyle\quad
+
\sum_{k=0}^{N-1}
\Big[
w_u\|\boldsymbol u_k\|^2
+
w_{\mathrm{ub}}
\big(
\|\max(\boldsymbol 0,\boldsymbol u_{\min}-\boldsymbol u_k)\|^2
+
\|\max(\boldsymbol 0,\boldsymbol u_k-\boldsymbol u_{\max})\|^2
\big)
\Big]
\\
&\text{s.t.}\quad
\boldsymbol q_{\min}\le \boldsymbol q_k \le \boldsymbol q_{\max},
\qquad k=1,\ldots,N. \nonumber
\end{align*}
Here \(w_p=10\), \(w_v=20\), \(w_\omega=20\), \(w_\theta=10\), \(w_u=10^{-2}\), and \(w_{\mathrm{ub}}=5\times10^2\). The max terms are inactive when \(\boldsymbol u_k\) lies within its velocity limits, and become positive only when the lower or upper velocity bound is violated. Thus, velocity limits are encouraged through a soft quadratic penalty, while joint-position bounds are enforced as hard constraints.

\section{Experiments} \label{sec:experiments}
\subsection{Experimental Setup and Implementation} \label{sec:experimental_setup}

\begin{figure} [t]
    \centering
    \subfigure[] {
    \includegraphics[width=0.45\textwidth]{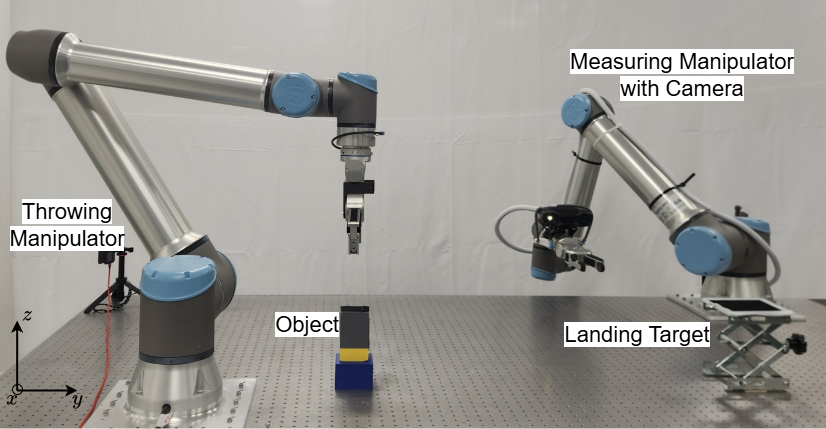}
    \label{fig:experiment_setup}
    }
    \subfigure[] {
    \includegraphics[width=0.4\textwidth]{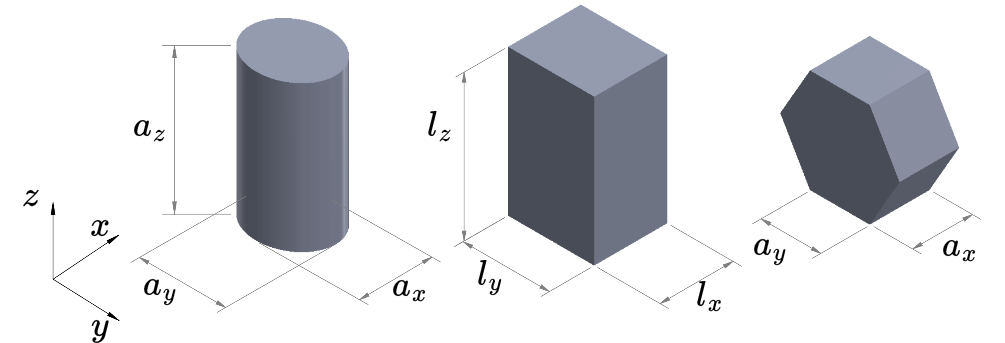}
    \label{fig:objects}
    }
    \vspace{-1em}
    \caption{
        (a) Experiment setup and  (b) objects used in the experiments and their geometric parameter definitions. The throwing motion is performed in the \(yz\)-plane.
    }
    \vspace{-1em}
\end{figure}

The experiments are conducted using a UR10e robot mounted on an optical table and equipped with an OnRobot RG6 gripper for grasping and releasing the objects, as depicted in Fig.~\ref{fig:experiment_setup}. During experiments, the trajectory generated by the swing generator, \(\{\boldsymbol q_k\}_{k=0}^{N}\), is converted into joint-velocity commands using finite differences, \(\dot{\boldsymbol q}_k \approx \frac{\boldsymbol q_{k+1} - \boldsymbol q_k}{h}\), and sent to the UR10e at a control frequency of \(500\,\mathrm{Hz}\), matching the trajectory discretization described in Swing-Trajectory Generator in Section~\ref{subsec:robot_swing_planner}.

Two parameters must be specified for the gripper command: the grasping force and the opening time. The nominal release is defined at the final trajectory index \(k=N\). The grasping force is selected to balance two competing effects: a larger force reduces slip risk during the swing, but slows the release transition, while a smaller force accelerates release but increases slip risk. To account for communication and actuation delays, the gripper-open command is issued before \(k=N\) so that the physical release occurs as close as possible to the nominal release time. The delay characterization is discussed in the following section.

We evaluate the framework on objects of varying shape, mass, and geometry, as shown in Fig.~\ref{fig:objects}. For each trial, the object properties---shape, mass, and geometric parameters---are provided to the planner before execution. The elliptical cylinder is included to test generalization across 3D geometries that share the same planar release model: its central cross-sectional dimensions match those of the block, so both objects use the same planar release-state solution. However, the cylinder's curved surface makes it harder to consistently grasp at the same location, unlike the block's flat faces.

At the start of each trial, the object is manually placed at a predefined picking location and the robot executes the same grasping motion; the object is aligned carefully to obtain a symmetric grasp, though small placement variations may remain. A landing table is positioned at distance $y_{\mathrm{des}}$ from the robot base and adjusted to height $z_{\mathrm{des}}$. The landing surface and objects are equipped with Velcro to suppress post-impact sliding. This aids the binary success criterion and may inflate success rates relative to smooth surfaces. Characterizing performance without this aid---or with surface materials representative of target applications---is a direction for future work.

To measure the landing pose, we use a high-resolution camera mounted on a second manipulator, enabling precise camera positioning. The camera-to-robot-base transformation is calibrated so that the measured object pose is expressed in the first robot's base frame. Each object is instrumented with an AprilTag on one face; after landing, the camera captures the AprilTag pose and the object center of mass (CoM) is recovered via the known rigid transformation from the AprilTag frame to the object CoM.

While the landing position can be measured accurately using the calibrated camera system, the task objective is stable landing on a prescribed support face rather than continuous orientation tracking. Accordingly, landing orientation is evaluated through a binary success criterion based on whether the object settles on the intended face and remains stable after impact.

\subsection{Evaluation Metrics}

A throw-flip attempt is counted as successful if the object lands stably on the intended face with the desired orientation. For successful trials, we compute the landing-distance error
\(
e_y = y_{\mathrm{land}} - y_{\mathrm{des}},
\)
where $y_{\mathrm{land}}$ is the measured CoM landing position along the $y$-axis, and report its mean and standard deviation.

We also analyze release timing to assess the instantaneous release assumption of our framework. Due to space limitations, we focus on this aspect alone, and approximate the effective release time using two timing estimates:
\begin{itemize}
    \item \textit{Gripper-motion delay}: the delay between the nominal release instant $k=N$ and the onset of gripper opening, detected as the first time the gripper width increases by at least $0.5~\mathrm{mm}$ above its initial value (chosen to exceed sensor noise while remaining sensitive to early motion).
    \item \textit{Detach-width delay}: the delay between $k=N$ and the time the gripper width crosses the detachment threshold---the width at which the grasp transitions from holding to releasing the object. This is estimated by commanding the gripper to close at minimum force until contact stops its motion.
\end{itemize}

We use the average of these two delays as the effective release time. A small gap between them indicates a near-instantaneous transition from gripper motion to object detachment.

\subsection{Experiment Results}

Across all experiments, we conduct 120 throwing trials, of which 108 are successful, giving an overall throw-flip success rate of \(90\%\). Results differ across objects: among the block-shaped objects, Block 1 achieves a slightly lower success rate and larger landing-distance error than Block 2, as shown in Table~\ref{tab:object_scenario_results}.

\begin{table}
\centering
\caption{Object properties, desired landing poses, and throwing results. \(\Delta t_{\mathrm{gap}}=\bar{t}_{\mathrm{detach}}-\bar{t}_{\mathrm{motion}}\) is the difference between the mean detach-width delay and the mean gripper-motion delay; the reported \(\bar{t}_{\mathrm{eff}}\) is the mean effective release delay normalized by the trajectory duration.}
\label{tab:object_scenario_results}
\resizebox{\textwidth}{!}{%
\begin{tabular}{c c c c c c c c c c}
\hline
Object & Mass \(m\) & Dimensions \(\boldsymbol d\)
& \(y_{\mathrm{des}}\) & \(z_{\mathrm{des}}\) & \(\theta_{\mathrm{des}}\)
& Succ. & \(\Delta t_{\mathrm{gap}}\) 
& \makecell{Normalized\\\(\bar{t}_{\mathrm{eff}}\)\\\((\times 10^{-3})\)}
& Mean / Std. \(e_y\) \\
 & (g) &  & (cm) & (cm) & (deg) &  & (ms) &  & (cm) \\
\hline

\multirow{3}{*}{Block 1}
& \multirow{3}{*}{100}
& \multirow{3}{*}{\makecell[l]{$(l_x,l_y,l_z)$\\$=(4,\,6,\,16)$ cm}}
& 120 & 10 & 180 & 8/10 & 22.82 & 7.15 & \(-5.178/1.210\) \\
& & & 127.5 & 15 & 180 & 9/10 & 20.20 & 1.92 & \(-3.278/1.411\) \\
& & & 135 & 20 & 180 & 10/10 & 20.30 & 21.03 & \(-0.792/1.647\) \\
\hline

\multirow{3}{*}{Block 2}
& \multirow{3}{*}{150}
& \multirow{3}{*}{\makecell[l]{$(l_x,l_y,l_z)$\\$=(5.5,\,6.5,\,11)$ cm}}
& 122.5 & 12.5 & 180 & 10/10 & 1.20 & 8.25 & \(-1.330/1.093\) \\
& & & 130 & 20 & 180 & 10/10 & 1.28 & 3.81 & \(-0.200/1.069\) \\
& & & 137.5 & 17.5 & 180 & 10/10 & 1.41 & 9.72 & \(1.287/1.065\) \\
\hline

\multirow{3}{*}{\makecell[c]{Elliptical\\Cylinder}}
& \multirow{3}{*}{150}
& \multirow{3}{*}{\makecell[l]{$(a_x,a_y,h_z)$\\$=(5.5,\,6.5,\,11)$ cm}}
& 122.5 & 12.5 & 180 & 8/10 & 1.22 & 13.77 & \(1.918/2.047\) \\
& & & 130 & 20 & 180 & 7/10 & 2.64 & 22.67 & \(1.634/1.575\) \\
& & & 137.5 & 17.5 & 180 & 8/10 & 11.29 & 13.24 & \(0.508/1.532\) \\
\hline

\multirow{3}{*}{\makecell[c]{Hexagonal\\Prism}}
& \multirow{3}{*}{70}
& \multirow{3}{*}{\makecell[l]{$(a_{\text{hex}},l_x)$\\$=(4.5,\,4.5)$ cm}}
& 125 & 17.5 & 60 & 10/10 & 1.69 & -6.70 & \(0.206/0.890\) \\
& & & 130 & 20 & 120 & 9/10 & -1.91 & -12.11 & \(-0.489/0.633\) \\
& & & 135 & 15 & 180 & 9/10 & 1.69 & -10.70 & \(-2.621/0.620\) \\
\hline
\end{tabular}
}
\vspace{-1em}
\end{table}

A likely contributing factor is the less instantaneous release observed for Block 1. Although the average effective release time remains close to the intended release instant, the gap $\Delta t_{\mathrm{gap}}$ between the gripper-motion and detach-width delays is substantially larger for Block 1 than for Block 2, indicating a slower release transition. This may introduce additional uncertainty in the realized release state, contributing to both the lower success rate and the larger landing-distance error. This slower transition is likely related to the higher grasping force required for Block 1: applying the lower force used for Block 2 causes Block 1 to slip during the swing. A secondary factor may be landing-face size: Block 2 presents a larger effective landing face, making stable post-impact landing easier. The performance gap between Block 1 and Block 2 may therefore reflect both release-timing uncertainty and differences in landing stability. A similar trend holds for the hexagonal prism, where a sharper release transition coincides with a high success rate and small mean landing-distance error. A final factor may be that Block 1 is the thinnest object in our object set, making its release more sensitive to nonlinearities in the gripper-opening speed.

A secondary pattern is the consistent negative bias in mean $e_y$ across all three Block 1 scenarios ($-5.178$, $-3.278$, and $-0.792\,\mathrm{cm}$), indicating that the object systematically lands short of the target. The bias diminishes as $y_{\mathrm{des}}$ increases, but since the traveled flight distance is comparable, it likely reflects unmodeled release dynamics rather than a fixed measurement offset.

For the elliptical cylinder, the release transition is also relatively sharp, but its curved surface reduces the repeatability of symmetric grasping with flat fingertips, potentially introducing small initial tilts prior to the swing. This is reflected in the data: success rates reach only 8/10, 7/10, and 8/10, and the standard deviation of $e_y$---2.047, 1.575, and 1.532\,cm---is roughly twice that of Block~2 under comparable targets. The worst-performing scenario ($y_{\mathrm{des}}=130\,\mathrm{cm}$, 7/10) also exhibits the largest effective release delay ($\bar{t}_{\mathrm{eff}}=22.67\times10^{-3}$), suggesting that timing variability compounds the grasp-consistency issue at this distance. Additionally, $\Delta t_{\mathrm{gap}}$ rises sharply to 11.29\,ms at $y_{\mathrm{des}}=137.5\,\mathrm{cm}$ compared with 1.22\,ms at $y_{\mathrm{des}}=122.5\,\mathrm{cm}$, indicating less consistent gripper actuation for this object. Despite these challenges, the cylinder achieves a moderate success rate across all scenarios, indicating that the planned trajectories transfer reasonably well to objects with curved geometry. 

Across the more favorable objects---Block~2 and the hexagonal prism---the timing metrics exhibit a consistent relationship with performance. For Block~2, all three scenarios achieve perfect success (10/10); the influence of release timing manifests instead in landing precision: the scenario with the smallest effective release delay ($\bar{t}_{\mathrm{eff}}=3.81\times10^{-3}$ at $y_{\mathrm{des}}=130\,\mathrm{cm}$) yields the lowest mean landing error ($|e_y|=0.200\,\mathrm{cm}$), while the two scenarios with larger delays ($\bar{t}_{\mathrm{eff}}=8.25\times10^{-3}$ and $9.72\times10^{-3}$) produce errors of 1.330 and 1.287\,cm, respectively. For the hexagonal prism, all mean $\bar{t}_{\mathrm{eff}}$ values are negative, indicating that the physical release consistently occurs slightly before the nominal instant; the scenario with the smallest deviation from nominal ($y_{\mathrm{des}}=125\,\mathrm{cm}$, $\bar{t}_{\mathrm{eff}}=-6.70\times10^{-3}$) is the only one to achieve perfect success (10/10) and the lowest landing error ($|e_y|=0.206\,\mathrm{cm}$), while both remaining scenarios ($\bar{t}_{\mathrm{eff}}=-12.11\times10^{-3}$ and $-10.70\times10^{-3}$) each incur one failure.

\subsection{Ablations}

\paragraph{Swing-pose selector.}
We evaluate the contribution of the swing-pose selector by replacing the optimized initial joint configuration with a fixed, hand-designed throwing posture similar to those in prior work~\cite{ref-tossingbot,ref-learning-throw-flip}. The ablation is performed in simulation across all 9 unique trajectories from the 12 scenarios in Table~\ref{tab:object_scenario_results}, where Block 2 and the elliptical cylinder share the same planned trajectory, keeping the desired release state unchanged and varying only the initial configuration provided to the swing-trajectory generator. The study is conducted in simulation to isolate the effect of initial-pose selection on release-state tracking, eliminating confounding factors such as release-timing uncertainty, grasp variability, and impact dynamics.

\begin{figure} [t]
    \centering
    \includegraphics[width=0.8\textwidth]{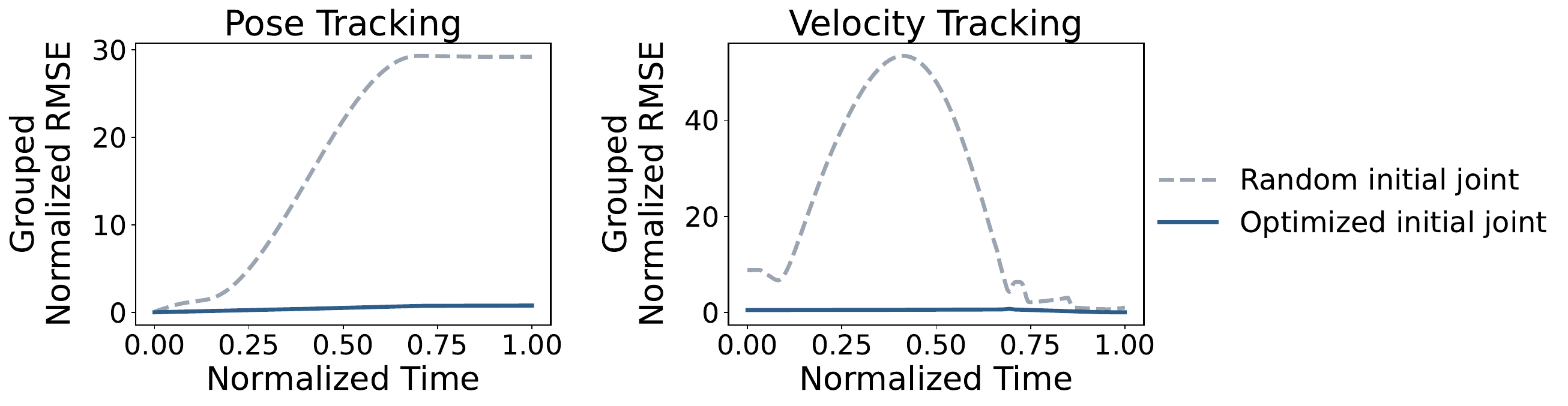}
    \vspace{-1em}
    \caption{
        Ablation of initial-pose selection.
        Grouped normalized pose- and velocity-tracking RMSE over normalized swing time, aggregated across all 9 unique trajectories.
    }
    \label{fig:ablation_initial_pose}
    \vspace{-.5em}
\end{figure}

Figure~\ref{fig:ablation_initial_pose} compares the resulting tracking errors. The optimized initial configuration consistently achieves lower normalized pose- and velocity-tracking RMSE over normalized swing time. The grouped errors are defined as $e_{\mathrm{pose}}(t_k)=
\|[\frac{\Delta p_y}{s_p},\, \frac{\Delta p_z}{s_p},\, \frac{\Delta \theta}{s_R} ]\|_2$ and $e_{\mathrm{vel}}(t_k)= \| [ \frac{\Delta v_y}{s_v},\, \frac{\Delta v_z}{s_v},\, \frac{\Delta \omega_x}{s_\omega}]\|_2$,
where $\Delta q = q - q^{\mathrm{ref}}$ denotes the tracking error of quantity $q$, and the normalization coefficients are the same as those used in Sec.~\ref{subsubsec: initial_joint_config_optimizer}. These results confirm that the swing-pose selector enables the robot to more accurately realize the desired release state.

\paragraph{Near-release constant-velocity interval.}
The decomposed framework also enables a structured near-release motion in which the end-effector's linear and angular velocities are held approximately constant immediately before release. This behavior is governed by \(\alpha\) in the Swing-Posture Selector: the adopted value \(\alpha=0.7\) reserves a dedicated constant-velocity window, whereas \(\alpha=1.0\) eliminates it entirely. Because the two settings may yield different release states, initial configurations, and swing trajectories---while achieving comparable simulated tracking error relative to their respective reference targets---this ablation specifically probes whether the structured near-release motion improves real-world robustness under non-ideal execution, a benefit not captured by simulation alone. Due to space constraints, we report two representative scenarios: Block 1 at \(y_{\mathrm{des}}=120\,\mathrm{cm}\), the lowest-performing Block~1 case with a less instantaneous release, and Block 2 at \(y_{\mathrm{des}}=130\,\mathrm{cm}\), the best-performing Block 2 case characterized by a more impulsive release. Together, these scenarios allow us to evaluate the effect of the near-release motion structure under both unfavorable and favorable release conditions.

\begin{table}
\centering
\caption{Ablation on the near-release motion structure. The nominal setting uses \(\alpha=0.7\), while the ablated setting uses \(\alpha=1.0\). }
\label{tab:ablation_near_release}
\footnotesize
\setlength{\tabcolsep}{3pt}
\renewcommand{\arraystretch}{1.15}
\begin{tabular}{c c c c c c c c}
\hline
Object & \(y_{\mathrm{des}}\) & \(z_{\mathrm{des}}\) & \(\theta_{\mathrm{des}}\)
& \multicolumn{2}{c}{\(\alpha=0.7\) (nominal)} 
& \multicolumn{2}{c}{\(\alpha=1.0\) (ablated)} \\
\cline{5-8}
 & (cm) & (cm) & (deg)
& Succ. & Mean / Std. \(e_y\) (cm)
& Succ. & Mean / Std. \(e_y\) (cm) \\
\hline
Block 1 & 120 & 10 & 180 & \(\mathbf{8/10}\) & \(\mathbf{-5.178}/1.210\) & 5/10 & \(-8.202/0.640\) \\
Block 2 & 130 & 20 & 180 & \(\mathbf{10/10}\) & \(\mathbf{-0.200}/1.069\) & 9/10 & \(-4.084/0.895\) \\
\hline
\end{tabular}
\vspace{-1em}
\end{table}

The proposed planner outperforms the ablated version on both scenarios, achieving an overall success rate of \(18/20\) versus \(14/20\) and consistently smaller mean landing errors. The improvement is particularly pronounced for Block 1, where release is less instantaneous, suggesting that the near-release constant-velocity structure reduces sensitivity to release-timing uncertainty. This behavior is consistent with the fact that a timing error $\Delta t$ perturbs the realized release velocity by approximately $a \Delta t$, where $a$ denotes the local end-effector acceleration. By reducing acceleration near release, the proposed structure improves robustness to timing uncertainty. In contrast, when \(\alpha=1.0\), the target release velocity is reached only at the end of the motion, causing even small release delays to produce larger deviations from the desired release state.

\section{Conclusion} \label{sec:conclusions}
We presented \emph{FlipItRight}, a decomposed framework for stable planar throw-flip across diverse objects and pose targets. Treating the release state as an explicit intermediate representation enables principled candidate filtering, adaptive swing-pose and initial-configuration selection, and structured near-release motion with approximately constant end-effector velocities. The framework achieves a $90\%$ success rate across 120 real-world trials and robustness beyond what simulation tracking error predicts. Limitations and future directions include extending to full spatial throw-flip with coupled rotational dynamics, characterizing sensitivity to object-property uncertainty, and replacing the empirical gripper pre-timing with an explicit model of non-instantaneous detachment.

\vspace{1em}
\noindent \textbf{Acknowledgments.} The work was supported by funding from King Abdullah University of Science and Technology (KAUST).
\vspace{-1em}

\bibliographystyle{spmpsci_unsrt}
\bibliography{refs}
\end{document}